\documentclass[11pt]{article}

\usepackage[hyperref]{ccl2025-en}
\hypersetup{hypertexnames=false}
\makeatletter
\let\cclOriginalLbibitem\@lbibitem
\def\@lbibitem[#1]#2{%
  \cclOriginalLbibitem[#1]{#2}%
  \Hy@raisedlink{\hyper@anchorstart{cite.#2}\hyper@anchorend}}
\makeatother
\usepackage{times}
\usepackage{url}
\usepackage{latexsym}
\usepackage{fancyhdr}
\usepackage[T1]{fontenc}
\usepackage[utf8]{inputenc}
\usepackage{microtype}
\usepackage{inconsolata}
\usepackage{graphicx}
\usepackage{flafter}
\usepackage{booktabs}
\usepackage{amsmath}
\usepackage{amssymb}
\usepackage{xcolor}
\usepackage{multirow}
\usepackage{tikz}
\usepackage{placeins}
\usetikzlibrary{arrows.meta,positioning}

\graphicspath{{figures/}}

\newcommand{\method}{UO-FIE}
\newcommand{\score}{\operatorname{Score}}

\title{UO-FIE: Combining Exact-Label Supervision with\\
Graded Utility for Factivity Inference}

\author{Xinchen Xiao \\
  Xinjiang University \\
  \texttt{xiaoxinchenn@outlook.com}}
\date{}

\begin{document}
\maketitle

\begin{abstract}
The Factivity Inference Evaluation 2026 (FIE2026) classifies Chinese context--hypothesis pairs into nine ordered factivity intervals. Its evaluation metric rewards both exact predictions and proximity to the correct interval, while 64.1\% of the 566 training examples belong to a single class. In preliminary experiments, several mDeBERTa classification models predominantly predict the dominant class, whereas a Huber-regression baseline produces more predictions near the correct interval but fewer exact matches.

We introduce Utility-Oriented Factivity Inference (\method), a parameter-efficient system that combines exact-label supervision with graded utility. \method{} predicts a distribution over the nine classes and combines hard-label supervision, utility-based soft targets, scheduled class weights, and an ordinal loss. We evaluate expected-utility decoding in controlled comparisons and use ordinal calibration selected on out-of-fold predictions for the submitted system.

Based on Qwen3.5-9B with LoRA, \method{} ranks first in the fine-tuning track with a macro utility of 0.8316. A separate prompt-based ensemble ranks third in the non-fine-tuning track with a macro utility of 0.8450.
\englishkeywords{factivity inference, ordinal classification, graded utility, utility-aligned learning, parameter-efficient fine-tuning}
\end{abstract}

\section{Introduction}

Factivity inference asks whether a linguistic context commits to the truth of an event, often through factive or counter-factive predicates, negation, modality, and reported speech. Prior work includes FactBank \cite{sauri-pustejovsky-2009-factbank} and neural models of lexicosyntactic inference \cite{white-etal-2018-lexicosyntactic}. FIE extends the problem to Chinese and asks systems to express both truth status and judgment strength \cite{cong-etal-2025-overview}.

FIE2026 adds a distance-sensitive graded utility to this linguistic problem \cite{fie2026-task}. Predictions occupy one of nine intervals, from strong counter-factive to strong factive. Exact intervals matter, but errors at distance one or two retain partial credit; more distant errors receive zero. Ordinary nine-class accuracy ignores this graded structure, whereas a smooth scalar objective can reduce exact-label accuracy. The 566 training labels are also sharply skewed: 363 occupy the strongest factive interval, while several intermediate intervals have fewer than ten examples.

Preliminary experiments reveal two recurring empirical patterns (Table~\ref{tab:baselines}). Because the dominant label and its neighborhood cover much of the data, a majority predictor already obtains 0.7652 micro utility, and several preliminary categorical mDeBERTa runs converge to the same pattern. Huber regression raises the within-two rate, the fraction of predictions at distance at most two, from 0.7827 to 0.8145, but lowers exact accuracy from 0.6413 to 0.1325. The first pattern predicts mainly the majority interval; the second produces more nearby predictions but fewer exact matches.

These observations favor a categorical model that preserves all nine labels while accounting for the distances between them. \method{} uses the official score matrix for this purpose and adapts Qwen3.5-9B with LoRA \cite{hu-etal-2021-lora}. The backbone estimates a nine-label distribution; an OOF-calibrated ordinal layer sets decision thresholds to account for class imbalance and task utility.

We report systems for both official resource settings. \method{} is the primary fine-tuning method, while prompt fusion provides an independent secondary submission.

We make three contributions:
\begin{itemize}
    \item We characterize two empirical patterns: majority-label collapse and regression toward nearby but inexact intervals.
    \item We introduce score-matrix-guided categorical learning that retains exact-label supervision while representing neighborhood utility and ordinal structure.
    \item We provide controlled OOF evidence for utility-aware training and decoder selection; the complete system ranks first in the official fine-tuning track.
\end{itemize}

\section{Related Work}

Event factuality models infer commitment from lexical, syntactic, and discourse evidence \cite{sauri-pustejovsky-2009-factbank,white-etal-2018-lexicosyntactic,zhang-etal-2023-document-factuality}. FIE2025 established the Chinese shared task \cite{cong-etal-2025-overview}, with systems based on fine-tuning, ensembling, and prompt arbitration \cite{gu-etal-2025-system,liu-etal-2025-system-report}. Other CCL evaluation reports have explored retrieval-augmented inference and multi-round voting for fine-grained Chinese prediction \cite{wang-etal-2025-srag-mav}. FIE2026 adds confidence-bearing intervals and banded utility \cite{fie2026-task}.

Cost-sensitive classification represents unequal mistakes with a cost matrix \cite{elkan-2001-cost-sensitive}; ordinal and label-distribution methods encode rank or neighborhood structure \cite{cao-etal-2020-ordinal,geng-2016-label-distribution,diaz-marathe-2019-soft-labels}; and decision theory selects actions under application loss \cite{gneiting-2011-point-forecasts}. \method{} combines these perspectives through FIE's score matrix.

\section{Task Formulation}

For a Chinese context--hypothesis pair $x=(c,h)$, a system predicts a factivity category and confidence. Evaluation maps the gold and predicted judgments to ordered values $y,\hat{y}\in\{0,\ldots,8\}$, which we call \emph{interval labels}. Labels 0--3 represent four counter-factive strengths, label 4 is non-factive, and labels 5--8 represent four factive strengths.

Let $d=|y-\hat{y}|$. The instance utility is
\begin{equation}
\score(y,\hat{y}) =
\begin{cases}
1.0000 & d=0,\\
0.9545 & d=1,\\
0.6827 & d=2,\\
0 & d\geq 3.
\end{cases}
\label{eq:score}
\end{equation}
Equation~(\ref{eq:score}) induces a symmetric score matrix $S$, where $S_{i,j}$ is the utility of predicting interval $j$ for gold interval $i$. The official ranking metric first averages utility within each gold class and then averages across the nine classes:
\begin{equation}
M_{\mathrm{macro}} = \frac{1}{9}\sum_{c=0}^{8}\frac{1}{N_c}
\sum_{n:y_n=c}S_{c,\hat y_n}.
\end{equation}
The organizer also reports an instance-level micro average as a reference statistic, but rankings are determined by $M_{\mathrm{macro}}$.
The diagonal of $S$ rewards exact predictions, the first two off-diagonal bands encode partial credit, and all more distant decisions receive zero. This matrix connects the task definition to \method{}.

\begin{figure}[t]
\centering
\includegraphics[width=\textwidth]{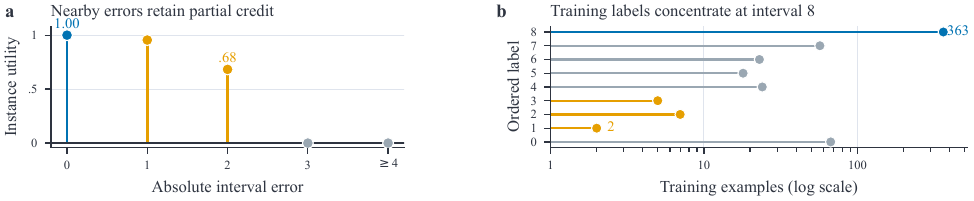}
\caption{The two structures that motivate our design. (a) Exact predictions receive full credit, nearby errors retain graded utility, and errors at distance three or more receive zero. (b) The 566 training labels are highly imbalanced: interval 8 contains 363 examples, whereas several intermediate intervals contain fewer than ten.}
\label{fig:task-geometry}
\end{figure}

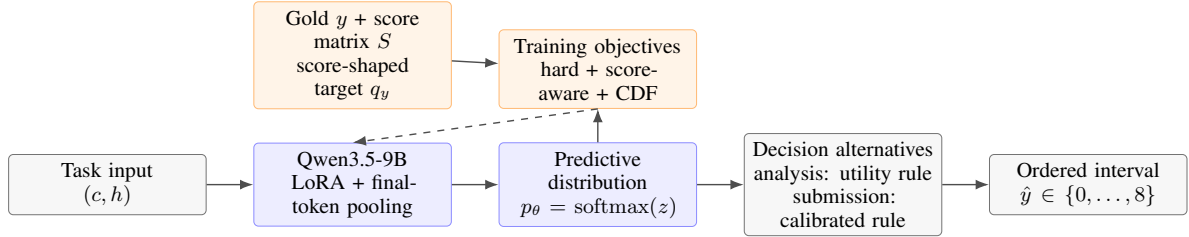
\begin{figure}[t]
\centering
\resizebox{0.98\textwidth}{!}{%
\begin{tikzpicture}[
  node distance=5mm and 7mm,
  box/.style={draw=black!55, fill=black!3, rounded corners=2pt, align=center,
              minimum height=9mm, text width=27mm, font=\small},
  key/.style={box, fill=blue!7, draw=blue!45},
  train/.style={box, fill=orange!9, draw=orange!55},
  arrow/.style={-{Latex[length=2mm]}, semithick, draw=black!70}
]
\node[box] (input) {Task input\\$(c,h)$};
\node[key, right=of input] (encoder) {Qwen3.5-9B\\LoRA + final-token pooling};
\node[key, right=of encoder] (posterior) {Predictive distribution\\$p_\theta=\mathrm{softmax}(z)$};
\node[box, right=of posterior] (decode) {Decision alternatives\\analysis: utility rule\\submission: calibrated rule};
\node[box, right=of decode] (output) {Ordered interval\\$\hat y\in\{0,\ldots,8\}$};
\node[train, above=of encoder] (target) {Gold $y$ + score matrix $S$\\score-shaped target $q_y$};
\node[train, above=of posterior] (loss) {Training objectives\\hard + score-aware + CDF};
\draw[arrow] (input) -- (encoder);
\draw[arrow] (encoder) -- (posterior);
\draw[arrow] (posterior) -- (decode);
\draw[arrow] (decode) -- (output);
\draw[arrow] (target) -- (loss);
\draw[arrow] (posterior.north) -- (loss.south);
\draw[arrow, dashed] (loss.south) -- (encoder.north);
\end{tikzpicture}
}
\caption{Overview of \method. Hard supervision and the score matrix shape a categorical distribution over exact labels and graded alternatives. Two decision alternatives operate on the same distribution: analysis uses expected utility, while the official submission uses prior correction and utility-selected ordinal calibration.}
\label{fig:overview}
\end{figure}

\section{Model Description: UO-FIE}

Figure~\ref{fig:overview} summarizes the fine-tuning pipeline. Both decoders use the same predicted distribution: development comparisons use expected-utility decoding, and the submitted system uses ordinal calibration.

\subsection{Utility-Aligned Training}

\paragraph{Keeping exact labels.}
We serialize the nine-class rubric, context, and hypothesis into one discriminative input. The fields follow this fixed order and are separated by blank lines. Qwen3.5-9B is used as the backbone \cite{qwen-team-2026-qwen35}; the preceding Qwen3 family is described by \newcite{yang-etal-2025-qwen3}. After left truncation to 1,024 tokens, the final non-padding hidden state passes through a linear nine-class head. The resulting predictive distribution is $p_\theta=\operatorname{softmax}(z)$. Only the classifier and LoRA adapters are trained; the backbone parameters remain frozen.

\paragraph{Turning scores into neighborhoods.}
Let $S\in\mathbb{R}^{9\times9}$ contain the utilities from Equation~(\ref{eq:score}). For gold class $y$, we derive a soft target
\begin{equation}
q_{y,k}=\frac{\exp(S_{y,k}/T_s)}{\sum_j\exp(S_{y,j}/T_s)},
\label{eq:softtarget}
\end{equation}
where $T_s=0.18$ controls target sharpness. Unlike generic label smoothing, Equation~(\ref{eq:softtarget}) assigns mass according to the evaluation neighborhood encoded by $S$. For an interior label, the gold interval remains the unique mode (approximately 0.34), while each immediate neighbor receives approximately 0.27. The combined neighboring mass represents the metric's credited band, and the hard-label term encourages exact predictions.

\paragraph{Balancing identity and graded error.}
Hard cross entropy encourages prediction of the correct interval. Three score-aware terms then organize probability mass by neighborhood utility and ordinal position:
\begin{align}
\mathcal{L}_{\mathrm{hard}} &= -\log p_y,\\
\mathcal{L}_{\mathrm{soft}} &= -\sum_k q_{y,k}\log p_k,\\
\mathcal{L}_{\mathrm{util}} &= -\log\left(\max\left\{\sum_k p_k S_{y,k},\epsilon\right\}\right),\\
\mathcal{L}_{\mathrm{cdf}} &= \frac{1}{9}\sum_k
\operatorname{SmoothL1}\left(\sum_{j\leq k}p_j,\sum_{j\leq k}q_{y,j}\right).
\end{align}
Here $\epsilon=10^{-8}$ keeps the logarithm finite. Score-shaped cross entropy encourages the prediction to follow the neighborhoods in $S$. The utility term places mass on labels that earn task credit; the lower-weight CDF term compares cumulative mass along the ordinal axis, so larger shifts affect more thresholds \cite{cao-etal-2020-ordinal}.

Let $\tilde{w}_y$ denote the normalized full class weight and let $\gamma(e)$ be the class scale in Table~\ref{tab:schedule}. The scheduled sample weight is $w_y(e)=1+\gamma(e)(\tilde{w}_y-1)$, and the weighted objective is
\begin{equation}
\mathcal{L}=w_y(e)\left(\lambda_h\mathcal{L}_{\mathrm{hard}}
+\lambda_s\mathcal{L}_{\mathrm{soft}}
+\lambda_u\mathcal{L}_{\mathrm{util}}
+\lambda_c\mathcal{L}_{\mathrm{cdf}}\right),
\end{equation}
where every $\lambda$ follows the schedule in Appendix Table~\ref{tab:schedule}. Across training, $\lambda_u$ rises from 0.42 to 0.55 while $\lambda_h$ falls from 0.15 to 0.04, shifting the objective toward utility alignment. Exact-label information remains in both hard supervision and the score-shaped target, whose unique maximum is at $y$.

\paragraph{Gradually increasing rare-class weights.}
Scheduled weighting handles label frequency under the class-balanced ranking objective. For class count $n_k$, let $u_k=\operatorname{clip}(\sqrt{N/(9n_k)},0.5,3.0)$. We apply factors $f_4=1.40$, $f_3=f_5=1.25$, $f_2=f_6=1.10$, and $f_k=1$ otherwise, then define $\tilde w_k=9u_kf_k/\sum_j u_jf_j$. These modest factors emphasize the central boundary and taper across adjacent intermediate labels. Clipping precedes boundary emphasis, and the final weights are renormalized to unit mean before their gradual introduction across training.

\subsection{Inference and Submission Calibration}

Given $p_\theta$, the analytical expected-utility rule predicts
\begin{equation}
\hat{y}=\arg\max_j\sum_i p_{\theta,i} S_{i,j}.
\label{eq:decode}
\end{equation}
When $p_\theta$ is a calibrated posterior estimate, Equation~(\ref{eq:decode}) is the Bayes action under task utility $S$. It differs from argmax when probability mass spans neighboring intervals. With only 566 highly skewed labels, however, the learned posterior can inherit class-prior bias. The official decision layer therefore applies a compact three-stage calibration: it adjusts the class prior and posterior sharpness, projects the corrected distribution to an expected ordinal position, and maps that position through monotone thresholds. Its low-dimensional parameters are selected by macro utility on stratified OOF predictions and then held fixed. Appendix~\ref{sec:submission-decoder} gives the decision rule, while Figure~\ref{fig:ablation} isolates the expected-utility decoder before calibration.

\section{Experiments}

\subsection{Data and Protocol}
\label{sec:data-protocol}

FIE2026 provides 566 labeled training instances. Both tracks use the same 2,958 context--hypothesis pairs and official annotations for evaluation, but the organizer maintains separate leaderboards for the two resource settings.

The fine-tuned model uses Qwen3.5-9B with LoRA rank 32, learning rate $10^{-4}$, batch size 4 with two gradient-accumulation steps, three epochs, classifier dropout 0.05, and LoRA dropout 0. Training uses 8-bit paged AdamW and 58.2M trainable parameters (0.689\% of 8.45B), taking 6.6 minutes on a single NVIDIA H800 GPU.

For development and calibration, we average per-instance logits from five stratified folds under three split seeds. These OOF predictions support model selection, calibration, and error analysis. The cross-validation models and the final model share the same architecture, schedule, and nine-logit output. A controlled single-seed sweep selects $T_s=0.18$ by relative OOF macro utility and far-error rate (Appendix Figure~\ref{fig:temperature-sensitivity}). After selecting the configuration, the final LoRA model is retrained on all 566 examples and uses the fixed decoder. We use OOF results for model selection and report final performance on the official evaluation set.

\subsection{Baseline Comparison: Exact Matches and Nearby Predictions}

Table~\ref{tab:baselines} compares baseline accuracy for exact matches and nearby predictions. The majority baseline always predicts label 8. The two LinearSVM ensembles combine character TF--IDF, structural, and cue templates; the full-cue variant expands uncertainty, negation, evidence, modality, and factuality cues. The regression baseline fine-tunes mDeBERTa-v3 \cite{he-etal-2021-debertav3} with Huber loss. These early baselines use three-fold OOF predictions; the final system follows the repeated five-fold protocol in Section~\ref{sec:data-protocol}.

\begin{table}[!ht]
\centering
\small
\setlength{\tabcolsep}{3pt}
\begin{tabular}{lrrr}
\toprule
System & Micro utility & Exact & Within-2 \\
\midrule
Always label 8 & 0.7652 & 0.6413 & 0.7827 \\
Low-template linear ensemble & 0.8342 & 0.6908 & 0.8551 \\
Full-cue linear ensemble & 0.8560 & 0.6873 & 0.8781 \\
mDeBERTa Huber regression & 0.7734 & 0.1325 & 0.8145 \\
\bottomrule
\end{tabular}
\caption{Development results on the 566 labeled instances. Learned systems use three-fold OOF evaluation. Micro utility, Exact, and Within-2 are instance-level averages. Majority prediction preserves apparent utility through class frequency; scalar regression reduces distance but lowers exact-label accuracy.}
\label{tab:baselines}
\end{table}

The majority predictor reaches 0.7652 micro utility despite using only one label. In preliminary experiments, several categorical mDeBERTa objectives reproduce this shortcut. Huber regression reverses the error profile: within-two accuracy rises to 0.8145, but exact accuracy falls to 0.1325. The linear systems show that lexical and structural evidence can improve utility without reducing the task to a scalar. These results therefore favor a categorical model that also encodes distance.

\subsection{Controlled Configuration and Decoder Comparisons}

\begin{figure}[!ht]
\centering
\includegraphics[width=\textwidth]{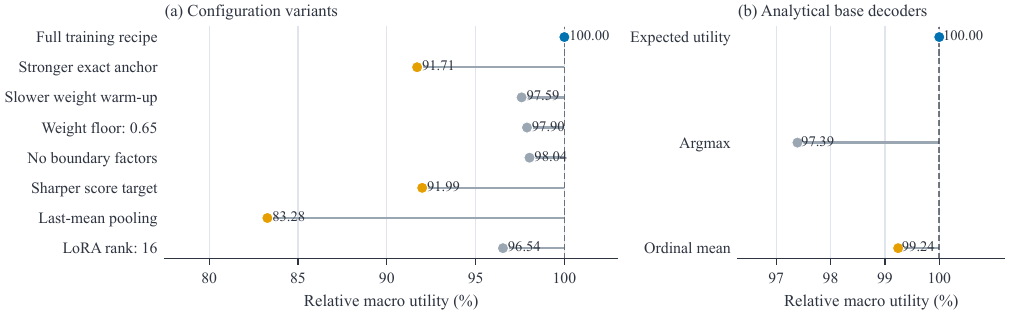}
\caption{Relative macro utility on repeated five-fold OOF predictions; higher is better. (a) Configuration variants use the same expected-utility decoder and are normalized by the full configuration. (b) Uncalibrated decoders use identical logits and are normalized by expected utility. Appendix~\ref{sec:submission-decoder} describes the submitted calibrated decoder.}
\label{fig:ablation}
\end{figure}

We compare fine-tuning configurations on repeated five-fold OOF predictions using macro utility, matching the official class-balanced aggregation. Figure~\ref{fig:ablation}(a) applies the same raw expected-utility decoder to every configuration. Figure~\ref{fig:ablation}(b) then holds logits fixed and changes only the uncalibrated decoder, separating decision effects from model changes.

Under the common decoder, warm-up, class-weight, boundary-factor, and lower-rank variants retain 96.54--98.04\% of the reference macro utility. Last-mean pooling, a sharper score target, and a larger hard-label loss weight retain 83.28--91.99\%. With unadjusted logits fixed, expected-utility decoding reaches 100.00\%, compared with 97.39\% for argmax and 99.24\% for ordinal-mean rounding. Appendix~\ref{sec:submission-decoder} defines the calibrated map used for the official submission.

\FloatBarrier
\subsection{Official Results}

As shown in Table~\ref{tab:official}, \method{} ranks first in the fine-tuning track, and five-prompt ordinal fusion ranks third in the prompt track.

\begin{table}[!ht]
\centering
\small
\setlength{\tabcolsep}{3pt}
\begin{tabular}{llrrr}
\toprule
Track & System & Rank & Macro avg. & Micro avg. \\
\midrule
Fine-tuning & \method{} LoRA & \textbf{1} & 0.831647 & 0.855997 \\
Prompt & Five-prompt ordinal fusion & 3 & 0.844968 & 0.886498 \\
\bottomrule
\end{tabular}
\caption{Official results on the same 2,958 pairs under separate resource settings. Rankings use macro utility; micro utility is shown for reference.}
\label{tab:official}
\end{table}

\subsection{Secondary Prompt-Track System}

The prompt system averages five 0--8 predictions from four semantic templates and two Gemini variants. All use the same rubric and three demonstrations, with different emphases on source attribution, factual commitment, scope, and boundaries between adjacent labels. The decoder assigns extreme labels only under unanimous agreement and preserves whether the mean prediction lies above or below the neutral label. Appendix~\ref{sec:appendix-prompt} gives the model configurations and decoding rule.

\FloatBarrier
\section{Analysis and Discussion}

The majority baseline's 0.7652 micro utility shows how an instance average can obscure sparse labels; scheduled weights and macro-utility analysis keep them visible. Post-hoc error analysis suggests that labels 2--6 and reported speech remain difficult (Appendix~\ref{sec:diagnostic-analysis}). Reported speech requires separating the writer's commitment from an attributed source, making intermediate boundaries and attribution clear targets for improvement.

Figure~\ref{fig:ablation}(b) compares analytical rules on fixed logits; the official system separately calibrates prior skew, sharpness, and class boundaries.

\subsection{Limitations}

This study uses 566 labeled examples and a single backbone, and the temperature sweep uses one random seed. The calibration is specific to FIE2026; its applicability to other ordinal tasks remains to be evaluated.

\section{Conclusion}

\method{} ranks first in the fine-tuning track, while the independent prompt system ranks third in its track. The results support combining exact-label supervision with graded utility in both training and decoding.

\section*{Acknowledgements}

The author is grateful to two friends who wish to remain anonymous: one for generously providing the computing resources and infrastructure needed for the experiments, and the other for their continued care, encouragement, and emotional support throughout this work. Without their help and encouragement, this work would have been much more difficult to complete.

\appendix

\section*{\LARGE Appendix}

The appendix provides supplementary analyses and implementation details: Section A examines score-target temperature, Section B presents post-hoc error analyses, Section C describes the prompt configurations and aggregation rule, Section D specifies the submitted decoder, and Section E summarizes the training settings.

\section{Score-Target Temperature}

\begin{figure}[ht]
\centering
\includegraphics[width=\textwidth]{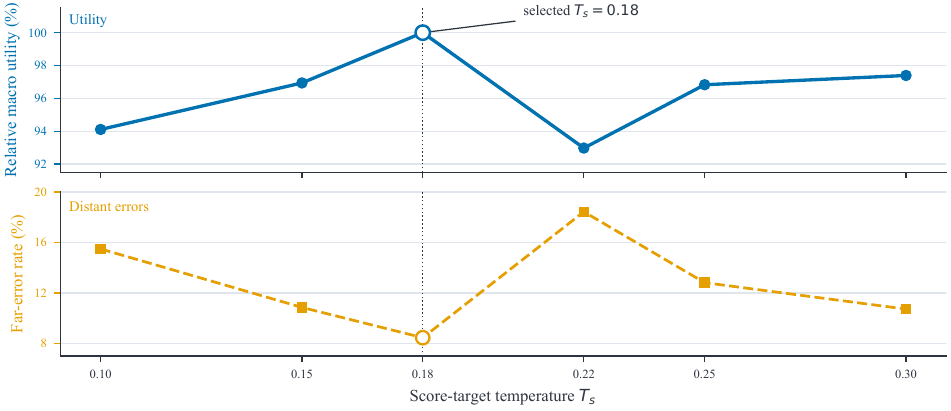}
\caption{Score-target-temperature sensitivity on stratified OOF development predictions. Relative macro utility is normalized to the selected $T_s=0.18$ run. The far-error rate counts predictions at least three intervals from the gold label. All points use one seed and the same expected-utility decoder.}
\label{fig:temperature-sensitivity}
\end{figure}

Figure~\ref{fig:temperature-sensitivity} reports the controlled sweep used to set $T_s$. All six runs share the same base configuration, fixed seed, and raw expected-utility decoder; only the score-target temperature changes. Relative macro utility peaks at 0.18, which we select for the final configuration.

\section{Post-hoc Error Analysis}
\label{sec:diagnostic-analysis}

\begin{figure}[ht]
\centering
\includegraphics[width=\textwidth]{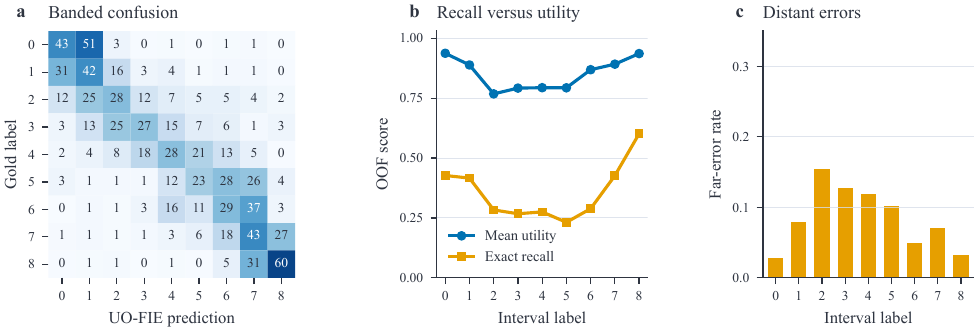}
\caption{Post-hoc error analysis using the separately constructed reference labels. (a) Row-normalized confusion percentages. (b) Exact recall and mean utility by class. (c) Far-error rate by class.}
\label{fig:diagnostic-errors}
\end{figure}

We analyze predictions from the submitted system using reference labels constructed separately for the shared evaluation inputs. These labels are used only for post-hoc analysis, not for system selection or official scoring. Figure~\ref{fig:diagnostic-errors} summarizes class behavior, with exact values in Table~\ref{tab:diagnostic-class}.

\begin{table}[ht]
\centering
\small
\setlength{\tabcolsep}{5pt}
\begin{tabular}{crr}
\toprule
Label & Recall & Utility \\
\midrule
0 & 0.427 & 0.937 \\
1 & 0.416 & 0.889 \\
2 & 0.284 & 0.768 \\
3 & 0.268 & 0.792 \\
4 & 0.275 & 0.794 \\
5 & 0.232 & 0.793 \\
6 & 0.289 & 0.869 \\
7 & 0.427 & 0.892 \\
8 & 0.601 & 0.936 \\
\bottomrule
\end{tabular}
\caption{Per-class post-hoc performance. Recall is exact-label recall; utility uses Equation~(\ref{eq:score}).}
\label{tab:diagnostic-class}
\end{table}

Labels 2--6 have the lowest exact recall, yet their mean utility remains moderate because nearby errors retain partial credit. This pattern suggests that intermediate boundaries remain harder than the outer intervals.

\begin{figure}[ht]
\centering
\includegraphics[width=\textwidth]{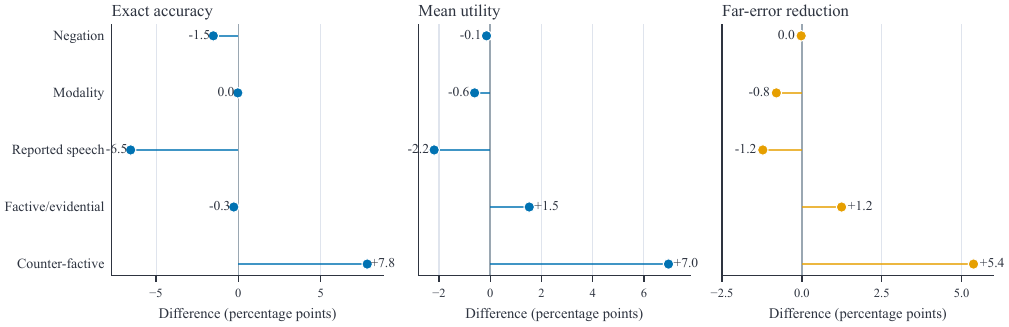}
\caption{Differences from overall post-hoc performance for overlapping subsets defined by lexical keywords. Positive values indicate improvement in all three panels. Reported speech is the weakest of the analyzed subsets.}
\label{fig:linguistic-slices}
\end{figure}

\FloatBarrier
\section{Prompt Configurations}
\label{sec:appendix-prompt}

We denote the five prompt configurations as R1--R5 according to their semantic roles.

\begin{center}
\centering
\small
\setlength{\tabcolsep}{4pt}
\begin{tabular}{@{}lll@{}}
\toprule
Configuration & Prompt emphasis & Model variant \\
\midrule
R1 & Comprehensive boundary reasoning & Gemini-3.5-Flash \\
R2 & Precision-oriented fact closure & Gemini-3.5-Flash \\
R3 & Balanced evidential reasoning & Gemini-3.5-Flash \\
R4 & Error-focused scope analysis & Gemini-3.5-Flash \\
R5 & Comprehensive boundary reasoning & Gemini-3.1-Pro-Preview \\
\bottomrule
\end{tabular}
\end{center}

\begin{figure}[ht]
\centering
\includegraphics[width=\textwidth]{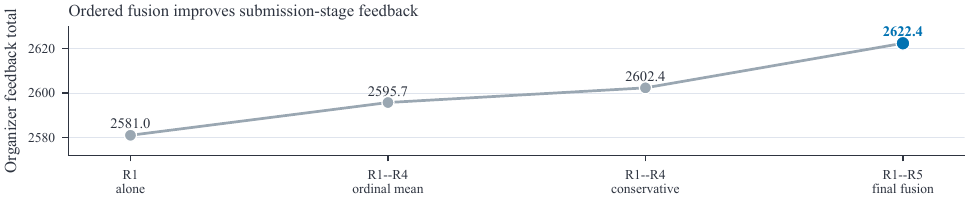}
\caption{Prompt-system evolution on the 2,958 shared inputs. The online submission interface returned an aggregate score for each stage; Table~\ref{tab:official} separately reports the final leaderboard's macro- and micro-average fields.}
\label{fig:prompt-evolution}
\end{figure}

Figure~\ref{fig:prompt-evolution} summarizes organizer feedback for R1 and the successive fusion stages, showing the progression from a single configuration to the final ensemble.

All configurations use temperature 0 and structured JSON output. R5 returned 2,955 of 2,958 responses; the three missing outputs were filled by R1, which uses the same prompt with Gemini-3.5-Flash. Let $a_r\in\{0,\ldots,8\}$ be the interval label from configuration $r$ and $m=\frac{1}{5}\sum_{r=1}^{5}a_r$. The fixed prompt decoding rule is
\begin{equation}
\hat y =
\begin{cases}
0, & m=0,\\
1, & 0<m<1,\\
\lfloor m+0.5\rfloor, & 1\leq m<3.5,\\
3, & 3.5\leq m<4,\\
4, & m=4,\\
5, & 4<m<4.5,\\
\lfloor m+0.5\rfloor, & 4.5\leq m<7,\\
7, & 7\leq m<8,\\
8, & m=8.
\end{cases}
\label{eq:prompt-decoder}
\end{equation}
Round-half-up is used on $1\leq m<3.5$ and $4.5\leq m<7$. The thresholds around label 4 preserve whether the mean prediction is above or below the neutral label, while labels 0 and 8 require unanimity.

\section{Ordinal Calibration}
\label{sec:submission-decoder}

Equation~(\ref{eq:decode}) assumes that the predictive distribution is calibrated. The submitted decoder follows the decision-theoretic distinction between probability estimation and action selection \cite{gneiting-2011-point-forecasts}. To compensate for prior bias under sparse, imbalanced supervision, it applies a deterministic class-prior adjustment followed by temperature scaling and monotone ordinal calibration. Let $z_k$ be the \method{} logit for interval label $k$ and $\pi_k$ its frequency in the 566-instance training set. It computes
\begin{align}
\tilde p_k &={} \frac{\exp\left((z_k-\tau\log\pi_k)/T\right)}
{\sum_{j=0}^{8}\exp\left((z_j-\tau\log\pi_j)/T\right)},\\
r &={} \sum_{k=0}^{8} k\tilde p_k,\\
\hat y &={} \sum_{j=1}^{8}\mathbf{1}[r\geq b_j],
\end{align}
where $\tau$, $T$, and $\mathbf{b}$ are the prior-correction strength, calibration temperature, and ordered threshold vector. The intermediate $r$ is the expected ordinal position, and the ordered thresholds keep the final map monotone. Calibration proceeds in two stages on repeated stratified OOF logits: a coarse search selects $\tau$ and $T$, followed by coordinate updates of $\mathbf{b}$ under monotonicity, minimum-spacing, and class-coverage constraints. The final settings, rounded to two decimal places, are $\tau=0.75$, $T=0.50$, and
\[
\mathbf{b}=(1.12,1.68,2.95,3.29,4.17,4.95,5.93,6.72).
\]
The selected decoder is held fixed when applied to the model retrained on all labeled examples. Figure~\ref{fig:ablation}(b) reports the corresponding decoder comparison with fixed logits.

\section{Training Details}

\begin{table}[ht]
\centering
\small
\begin{tabular}{lrrrrr}
\toprule
Epoch & Hard & Soft & Utility & CDF & Class scale \\
\midrule
1 & 0.15 & 0.35 & 0.42 & 0.08 & $0\rightarrow0.5$ \\
2 & 0.08 & 0.32 & 0.48 & 0.12 & $0.5\rightarrow1$ \\
3 & 0.04 & 0.28 & 0.55 & 0.13 & $1$ \\
\bottomrule
\end{tabular}
\caption{Loss and class-weight schedule for the final fine-tuned model.}
\label{tab:schedule}
\end{table}

\begin{itemize}
    \setlength{\itemsep}{0pt}
    \item Data: five-fold OOF under three split seeds; final training on all 566 released labeled instances.
    \item Backbone: Qwen3.5-9B.
    \item Training: LoRA $r=32$, $\alpha=64$, adapter dropout 0, classifier dropout 0.05, three epochs, learning rate $10^{-4}$, effective batch size 8; a single NVIDIA H800 GPU, 6.6 minutes.
\end{itemize}

\bibliographystyle{ccl}
\bibliography{references}

\end{document}